\documentclass[conference]{IEEEtran}
\IEEEoverridecommandlockouts
\usepackage[left=0.625in, right=0.625in, top=0.75in, bottom=0.75in]{geometry} %==
\usepackage{float}  % In your preamble
\usepackage{algorithm}
\usepackage{algpseudocode}
\usepackage{cite}
\usepackage{amsmath,amssymb,amsfonts}
\usepackage{graphicx}
\usepackage{textcomp}
\usepackage{balance}
\usepackage{tabularx}
\usepackage{graphicx}
\usepackage{adjustbox}
\usepackage{amsmath}
\usepackage{booktabs}  % Optional: cleaner horizontal rules
\usepackage{multirow}
\usepackage{mathrsfs}
\usepackage{enumitem}
\usepackage{stfloats}  % In preamble
\usepackage{placeins} % For \FloatBarrier
\usepackage{url}
\usepackage{xcolor}

\usepackage{tikz}
\usepackage{amsmath,amssymb}
\usepackage{xcolor}
\usepackage{enumitem}
\usetikzlibrary{arrows.meta, positioning, calc, shapes.geometric,
                fit, backgrounds, decorations.pathmorphing}
 
\definecolor{stage0blue}{HTML}{3b7bb8}
\definecolor{stage1teal}{HTML}{2e9d88}
\definecolor{stage2orange}{HTML}{d98227}
\definecolor{stage3red}{HTML}{c0486a}
\definecolor{stage3green}{HTML}{00FF00}
\definecolor{stage4gray}{HTML}{6b6b7a}
\definecolor{safebg}{HTML}{fff7e6}
\definecolor{sidebarbg}{HTML}{fbfbfd}
\definecolor{winbg}{HTML}{d8efd8}
\definecolor{modeblue}{HTML}{1F6FB2}
\definecolor{gradgreen}{HTML}{2E9D50}

\definecolor{revgreen}{HTML}{0B7A0B}

\algrenewcommand\algorithmiccomment[1]{\hfill\makebox[2.5cm][l]{\(\triangleright\) #1}}

\begin{document}
\title{Towards Robust Reinforcement Learning for Small-Scale Language Model Agents}

\author{Md Rezwanul Haque\textsuperscript{1}, Md. Milon Islam\textsuperscript{2},  and Fakhri Karray\textsuperscript{1,3}

\thanks{\textsuperscript{1}The authors are with the Centre for Pattern Analysis and Machine Intelligence, Department of Electrical and Computer Engineering, University of Waterloo, N2L 3G1, Ontario, Canada. (e-mail: rezwan@uwaterloo.ca{*}).

\textsuperscript{2}The author is with the Department of Computer Science and Engineering, Khulna University of Engineering \& Technology, Khulna, Bangladesh. (e-mail: milonislam@cse.kuet.ac.bd, milonislam@uwaterloo.ca).

\textsuperscript{1,3}The author is with the Centre for Pattern Analysis and Machine Intelligence, Department of Electrical and Computer Engineering, University of Waterloo, N2L 3G1, Ontario, Canada, and Department of Machine Learning, Mohamed bin Zayed University of Artificial Intelligence, Abu Dhabi, United Arab Emirates. (e-mail: karray@uwaterloo.ca, fakhri.karray@mbzuai.ac.ae).}
}
\maketitle

% %%%%%%----------------------------------------------------------
% \begingroup
% \renewcommand\thefootnote{}

% \footnotetext{
% % \noindent\rule{0.97\linewidth}{0.2pt}\\[2pt]
% \textsuperscript{*}Correspondence to: Md Rezwanul Haque \texttt{<rezwan@uwaterloo.ca>}.
% }
% \footnotetext{
% \textsuperscript{©}\textit{Proceedings of the 2026 IEEE International Conference on Systems, Man, and Cybernetics (SMC), Bellevue, WA, USA. Copyright 2026 by the author(s).}
% }
% \endgroup
% %%%%%--------------------------------------------------------

% ─── Abstract ────────────────────────────────────────────────────────────────

\begin{abstract}
The alignment of Small Language Models (SLMs) in the 70--500M parameter range using reinforcement learning is often considered unstable, though the underlying failure mechanisms have not been systematically investigated. In the State-of-the-Art (SOTA) research, fifteen (model, corpus) configurations were trained using Proximal Policy Optimization (PPO). The experiments included Pythia-70M, 160M, 410M and SmolLM2-135M, 360M on the TinyStories, CNN/DailyMail, and Wikitext-103 corpora. Three reproducible failure modes were identified in small-scale language models: silent LoRA parameter freezing in standard PEFT/TRL pipelines, numerical overflow in importance ratios when using bfloat16, and catastrophic policy collapse due to reward-model error. These issues were addressed using a merge-and-reinitialize adapter technique, float32 precision during PPO updates, and a three-layer safety mechanism comprising reward whitening, importance-ratio guarding, and weight rollback. In this paper, a capacity-headroom hypothesis is proposed, which states that PPO performance at the SLM scale depends on both a fluent supervised model ($\text{PPL}<20$) and a discriminative reward signal, rather than on the number of model parameters. The proposed system converged stably in all experiments and improved preference win rate over the SFT baseline in configurations with a fluent prior and an informative reward signal. Furthermore, it outperformed instruction-tuned baselines while requiring significantly less training data. All checkpoints, preference datasets, and training scripts are publicly released\textsuperscript{\S}. %released\footnotemark.
\end{abstract}

%%%%%%-----------------------------------
\begingroup
\renewcommand{\thefootnote}{}
\footnotetext[1]{%
% \noindent\rule{0.96\linewidth}{0.2pt}\\[2pt]
\hspace*{-\footnotesep}\rule{\dimexpr0.965\linewidth+\footnotesep\relax}{0.2pt}\\[-1pt]
\scriptsize
\textsuperscript{\S}Source code: \url{https://github.com/rezwanh001/SLM-RL-Agents} \\
Model checkpoints: \url{https://huggingface.co/mr3haque/SLM-RL-Agents} \\
Preference datasets: \url{https://huggingface.co/datasets/mr3haque/SLM-RL-Agents-Data}
}
\endgroup

% \footnotetext{
% \scriptsize Source code: \url{https://github.com/rezwanh001/SLM-RL-Agents} \\ 
% Model checkpoints: \url{https://huggingface.co/mr3haque/SLM-RL-Agents} \\ 
% Preference datasets: \url{https://huggingface.co/datasets/mr3haque/SLM-RL-Agents-Data}}
%%%%%%-----------------------------------

%%%%%%---------This is for only ArXiv Version--------------
\begingroup
\renewcommand\thefootnote{}
\footnotetext{
% \noindent\rule{0.97\linewidth}{0.2pt}\\[2pt]
\hspace*{-\footnotesep}\rule{\dimexpr0.225\linewidth+\footnotesep\relax}{0.2pt}\\[-1pt]
\textsuperscript{*}Correspondence to: Md Rezwanul Haque \texttt{<rezwan@uwaterloo.ca>}.
}

\footnotetext{
\textsuperscript{©}\textit{Proceedings of the 2026 IEEE International Conference on Systems, Man, and Cybernetics (SMC), Bellevue, WA, USA. Copyright 2026 by the author(s).}
}
\endgroup
%%%%%%-----------------------------------

%% ─── End Abstract ────────────────────────────────────────────────────────────────


% ─── 1. Introduction ─────────────────────────────────────────────────────────
%\vspace{-1.6em}
\section{Introduction}
\label{sec:intro}

Small Language Models (SLMs) in the 70--500M parameter range are suitable for resource-constrained, privacy-preserving, and on-device agents in human--machine systems. Unlike billion-parameter models, SLMs can perform real-time inference on commonly available hardware, facilitating deployment in agentic systems for tasks such as summarization, controlled generation, and dialogue at the edge. The alignment of SLM agents via reinforcement learning is important for both machine learning and systems engineering, as the PPO control loop must remain stable at scales where the learning signal is weak and the policy capacity is limited. Models with more than 100 billion parameters have achieved human-level performance across various benchmarks~\cite{brown2020language}. However, high training and deployment costs make them difficult to use in academic laboratories and edge-deployed applications. Data curation and architectural improvements have reduced the performance gap between SLMs and larger models~\cite{eldan2023tinystories, biderman2023pythia, allal2025smollm2, liu2024mobilellm, qwen2025technical}, making models with fewer than 500 million parameters a practical choice for on-device inference.

%%% ==================================

This simplified formulation provides a stability foundation for more advanced applications involving tool use and multi-turn interactions.
However, the alignment of SLMs remains an open research problem. Reinforcement Learning from Human Feedback (RLHF)~\cite{christiano2017deep, ouyang2022training} is the standard method for aligning large-scale models, though application of the models with fewer than 500 million parameters remains limited. PPO~\cite{schulman2017proximal} is commonly considered unstable at this model scale, often leading to exploding gradients and reward collapse. Therefore, Supervised Fine-Tuning (SFT) or Direct Preference Optimization (DPO)~\cite{rafailov2023direct} is often used instead, without investigating the underlying causes of PPO instability. This assumption is used to determine whether classical PPO can effectively support small language model agents and identify the technical requirements for stable training. The agent is represented as a parameterized policy $\pi_\theta$~\cite{schulman2017proximal} that selects actions from a discrete action space and receives scalar reward from the environment. The fifteen configurations considered in this research use a single-turn form of the resulting Markov Decision Process (MDP). In this setting, the action space is defined by the vocabulary $\mathcal{V}$, the horizon $T$ consists of a single turn ending with the end-of-sequence token, and the reward is provided at the terminal state. The empirical evaluation is restricted to the single-turn case, while the framework for multi-turn interactions is released with this paper.

%%% ================================

This question is addressed through an empirical study using an end-to-end system. Five base models (Pythia-70M, 160M, 410M and SmolLM2-135M, 360M)~\cite{biderman2023pythia, allal2025smollm2} are trained on three distinct datasets (TinyStories~\cite{eldan2023tinystories}, CNN/DailyMail~\cite{nallapati2016abstractive}, and Wikitext-103~\cite{merity2016pointer}) using a full SFT $\to$ reward-model $\to$ PPO cycle. This process generates fifteen fully trained configurations. All hyperparameters are kept the same across all experiments to investigate the effects of model capacity and domain difficulty. This approach allows us to distinguish configuration-specific issues from general structural failure modes.

%%% ================================
Three engineering failure modes specific to the small-scale architectures were identified. First, in certain Parameter-Efficient Fine-Tuning (PEFT) implementations of the Transformer Reinforcement Learning (TRL) library~\cite{vonwerra2022trl}, Low-Rank Adaptation (LoRA) parameters are silently registered as non-trainable. Second, bfloat16 arithmetic generates probability-ratio overflow in models with fewer than 200 million parameters. Third, unbounded reward values combined with unclipped Kullback--Leibler (KL) penalties can cause the policy to generate incoherent outputs. These failure cases are addressed using a merge-and-reinitialize method, float32 precision during training, and a multi-layer safety framework incorporating reward whitening and a weight-rollback mechanism.

A \emph{capacity-headroom hypothesis} is proposed, suggesting that PPO effectiveness depends on both a fluent SFT prior and a discriminative reward signal, rather than on the model parameter count. The proposed system achieves performance comparable to publicly available instruction-tuned baselines, including SmolLM2-Instruct~\cite{allal2025smollm2} and Qwen2.5-Instruct~\cite{qwen2025technical}, while using significantly less training data.

%%%%% ============================= 

The major contributions of this work are as follows:
\begin{enumerate}[leftmargin=*, topsep=1pt, itemsep=1pt]
  \item Three reproducible failure modes of PPO at the SLM scale are identified: silent LoRA gradient freezing in PEFT, bfloat16 importance-ratio overflow, and reward-driven policy collapse. Specific solutions are proposed and organized as a three-layer cybernetic safety framework.
  \item The capacity-headroom hypothesis is evaluated across fifteen configurations. The results show that PPO effectiveness at the SLM scale depends on the combination of SFT fluency and reward discriminability, providing a decision rule ($\text{PPL}_{\text{SFT}}<20$) for practitioners.
  \item A reproducible alignment framework is released, including fifteen checkpoints, preference datasets, training scripts, an interactive verification application, and a forward-compatibility system for multi-turn agentic extensions.
\end{enumerate}

The remainder of this paper is organized as follows. Section~\ref{sec:related} reviews related works. Section~\ref{sec:method} describes the methodology and stabilization framework. Section~\ref{sec:setup} presents the experimental setup. Section~\ref{sec:results} presents and analyzes the experimental results. Section~\ref{sec:discussion} discusses practical implications and Section~\ref{sec:conclusion} concludes the paper.

% ─── End 1. Introduction ─────────────────────────────────────────────────────

% ─── 2. Related Work ─────────────────────────────────────────────────────────
%\vspace{-0.5em}
\section{Related Works}
\label{sec:related}
The application of RLHF using SFT, reward modeling, and PPO~\cite{christiano2017deep, stiennon2020learning, ouyang2022training, schulman2017proximal} is well developed for models with 1.3–175 billion parameters~\cite{touvron2023llama}. However, its behavior at the 70–500 million parameter scale has received limited attention. Existing SLM-alignment research primarily focuses on developing high-quality supervised models through SFT~\cite{eldan2023tinystories, biderman2023pythia, allal2025smollm2} or using preference-learning methods that avoid the explicit reinforcement-learning process, such as DPO~\cite{rafailov2023direct}, KTO~\cite{ethayarajh2024kto}, and GRPO~\cite{shao2024deepseekmath}. While these alternative methods reduce training instability, they remove the explicit reward model, which is important for diagnostic transparency in agentic systems. 

Parameter-efficient training using LoRA~\cite{hu2022lora} and QLoRA~\cite{dettmers2023qlora}, typically implemented through the TRL library~\cite{vonwerra2022trl}, is a standard approach. However, silent gradient-flow problems can occur at the small-scale model level. Similarly, stabilization techniques such as adaptive KL~\cite{ziegler2019fine}, Generalized Advantage Estimation (GAE)~\cite{schulman2015high}, and NEFTune~\cite{jiang2023neftune} are adopted from large-scale models without sufficient evaluation at the SLM scale. This work provides a different research direction from the DPO/GRPO-based approaches. This work identifies reproducible failure modes of standard PPO at the SLM scale, rather than proposing a new optimization objective, and provides engineering solutions to maintain the diagnostic value of a scalar reward signal for practitioners.

%%%%%──────────────────────────── End 2. Related Work───────────────────────────

% ─── 3. Methodology ──────────────────────────────────────────────────────────
\vspace{-0.5em} %% for ArXiv only
\section{Methodology}
\label{sec:method}

\begin{figure*}[t]
\centering
% ### L, B, R, T

% \includegraphics[width=\textwidth, trim=63mm 49mm 63mm 17mm, clip]{figure/SLM_PL-h-2.pdf} %%% stnd

% \includegraphics[width=\textwidth, trim=3mm 11mm 3mm 0mm, clip]{figure/RLHF_Pipeline 4x.png} %%% stnd

\includegraphics[width=\textwidth, trim=3mm 11mm 3mm 0mm, clip]{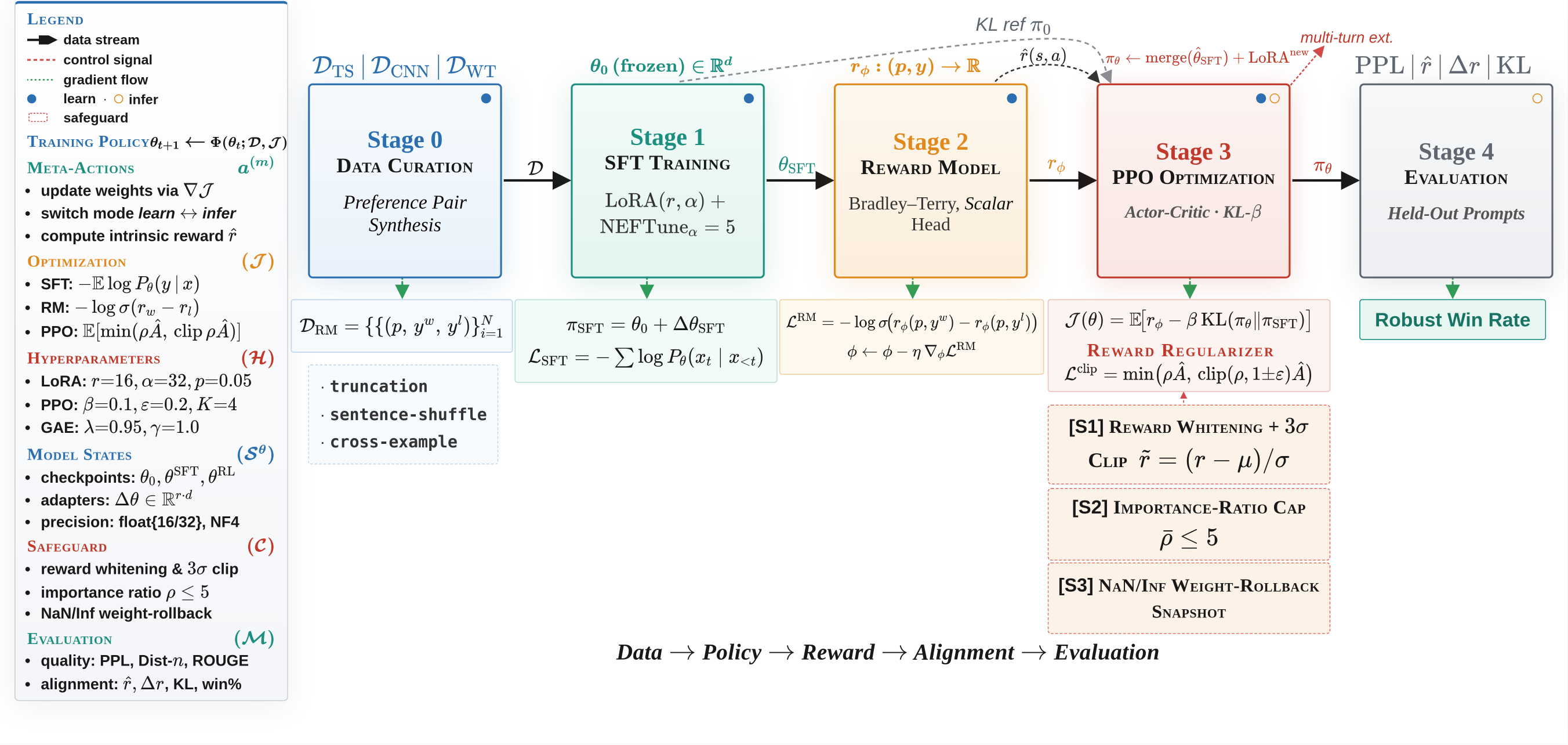}

% \includegraphics[width=\textwidth, trim=0mm 1mm 1mm 0.5mm, clip]{figure/SLM_PL.png} 
%\vspace{-0.9em}
% \caption{\textbf{End-to-end RLHF pipeline for aligning small language model agents.} The pipeline consists of five sequential stages: (0) Data Curation, where preference pairs are generated using truncation, shuffling, and mismatch; (1) SFT Training, generating the supervised prior $\pi_{\text{SFT}}$; (2) Reward Model training using the Bradley--Terry objective; (3) PPO Optimization, applying a merge-and-reinitialize procedure and a three-layer safety mechanism (reward whitening, importance-ratio capping, and weight-rollback) to reduce scale-specific instabilities; and (4) Evaluation using held-out prompts. Solid arrows represent primary data flow, dashed red arrows indicate control signals (e.g., KL reference), and dotted arrows show gradient flow into the state boxes of each stage. The colored dots represent operating mode (blue: learn; gold: inference).}
\vspace{-0.5em} %% for ArXiv only
\caption{\textbf{End-to-end RLHF pipeline for aligning small language model agents.} The pipeline consists of five sequential stages: (0) Data Curation, where preference pairs are generated using truncation, shuffling, and mismatch; (1) SFT Training, generating the supervised prior $\pi_{\text{SFT}}$; (2) Reward Model training using the Bradley--Terry objective; (3) PPO Optimization, applying a merge-and-reinitialize procedure and a three-layer safety mechanism (reward whitening, importance-ratio capping, and weight-rollback) to reduce scale-specific instabilities; and (4) Evaluation using held-out prompts. Solid arrows represent primary data flow, dashed red arrows indicate control signals (e.g., KL reference), and dotted arrows show gradient flow into the state boxes of each stage. The colored dots represent operating mode (blue: learn; gold: inference). \emph{Sidebar notation:} $\hat{r}(s,a)$ is the reward from the Bradley--Terry model $r_\phi(p,y)$, and $(s,a)$ reduces to $(p,y)$ in the single-turn setting; $\Phi$ and $\mathcal{J}$ denote the update operator and PPO objective $J(\theta)$, both distinct from the CDF $\Phi$ and MDP tuple $\mathcal{M}$ used in the text.}
\label{fig:pipeline}
\end{figure*}

A three-stage RLHF system is implemented, including supervised fine-tuning, Bradley--Terry reward modeling, and PPO with a KL divergence penalty. Each stage is tailored to models with fewer than 500 million parameters. The overall data flow and the three engineering mechanisms used to ensure training stability are illustrated in Fig.~\ref{fig:pipeline}.

%%%%-----------------------

%%%%%%%#################################################################%%%%%%%%

The agent policy $\pi_\theta$ operates within a finite-horizon MDP $\mathcal{M}=(\mathcal{S},\mathcal{A},\mathcal{P},\mathcal{R},T)$. The state $s_t\in\mathcal{S}$ is represented as a structured tuple containing the task prompt, generated tokens, external observations, the turn index, and the remaining budget. The action space $\mathcal{A}$ consists of four distinct action types: vocabulary token generation $a\in\mathcal{V}$, external tool invocation with arguments, clarifying-query generation, and termination. The transition function $\mathcal{P}$ is deterministic for vocabulary generation and termination, stochastic for tool invocation, and history-dependent in all cases. The reward $\mathcal{R}$ consists of three components: a verifiable component based on deterministic checks, a preference component obtained from the Bradley–Terry model $r_\phi$, and a shaping component based on the per-token KL divergence from the SFT prior.

The fifteen empirical configurations use the single-turn form of $\mathcal{M}$, where $T=1$, $\mathcal{A}=\mathcal{V}$, $\mathcal{P}$ is deterministic, the reward is provided only at the terminal state, and the observation set is empty. In this form, the state is reduced to the context $(x, y_{<t})$ and the MDP follows the standard language-modeling formulation used in large-scale alignment~\cite{stiennon2020learning, ouyang2022training}. The stability requirements investigated in this paper, including gradient-flow integrity, importance-ratio computation, and reward regulation, are necessary for any form of $\mathcal{M}$ and can also be applied to the multi-turn version released as a forward-compatibility layer. From a control-theoretic perspective, the PPO update can be viewed as a discrete-time feedback loop involving three connected signals: reward, KL divergence, and importance ratio. The safety mechanism described in Section~\ref{sec:ppo} provides specific controls that improve the robustness of the loop against reward-model misspecification. The three failure modes mentioned in Section~\ref{sec:ppo} are considered as closed-loop instabilities and are addressed at the controller level using signal conditioning, deadband limiting, and state recovery.

%\vspace{-0.6em}
% \subsection{Stage~0: Data Preparation and Preference Pair Construction}
\subsection{Stage~0: Data and Preference Pair Preparation}
%\vspace{-0.4em}
\label{sec:data}

Three target corpora, TinyStories ($\mathcal{D}_{\text{TS}}$), CNN/DailyMail ($\mathcal{D}_{\text{CNN}}$), and Wikitext-103 ($\mathcal{D}_{\text{WT}}$), are divided into separate splits for SFT training, preference-pair construction for reward model training, and held-out evaluation. Synthetic preference pairs are generated using three different degradation strategies. Truncation generates rejected continuations by cutting a sentence at 50\% of its length, usually in the middle of a clause, to penalize premature stopping. Sentence shuffling randomly changes the order of sentences in multi-sentence passages to penalize overall incoherence. Cross-example mismatch pairs a prompt with a reference text from another example to enforce topical coherence. These three strategies are used in equal proportions to construct the preference dataset $\mathcal{D}_{\text{RM}}=\{(p_i, y_i^{w}, y_i^{l})\}$.

%\vspace{-0.6em}
\subsection{Stage~1: Supervised Fine-Tuning}
%\vspace{-0.4em}
\label{sec:sft}

The causal language-modeling objective is minimized using the SFT dataset $\mathcal{D}_\text{SFT} = \{x^{(i)}\}_{i=1}^{N}$, as defined in~(\ref{eq:sft}).
\begin{equation}
  \mathcal{L}_\text{SFT}(\theta) \;=\; -\frac{1}{N}\sum_{i=1}^{N}\sum_{t} \log P_{\theta}\!\left(x^{(i)}_{t}\,\big|\,x^{(i)}_{<t}\right)
  \label{eq:sft}
\end{equation}
Here, $\theta_0\in\mathbb{R}^{d}$ represents the pre-trained base-model parameters, which are frozen. The updates are applied only to the LoRA adapter~\cite{hu2022lora}, parameterized by $(A\in\mathbb{R}^{r\times d}, B\in\mathbb{R}^{d\times r})$, where $r\in\{8,16,32\}$ depends on the model size and $\alpha=2r$. The effective weight matrix is given by $W_0 + \tfrac{\alpha}{r} BA$. LoRA is applied to all attention projection matrices and the dense layers of the Feed-Forward Network (FFN). In the GPT-NeoX architecture~\cite{gao2020pile}, this includes the combined query-key-value projection and the primary FFN up-projection. For Llama-style architectures \cite{touvron2023llama}, LoRA updates are applied to the individual query, key, and value projections, as well as the gate and down-projection matrices. NEFTune noise ($\alpha_{\text{NEFT}}=5$)~\cite{jiang2023neftune} is applied to reduce over-memorization during training. Optimization is performed using AdamW~\cite{loshchilov2017decoupled} at a peak learning rate of $2\times10^{-5}$, with a 6\% linear warm-up and cosine decay over 5 epochs. The resulting adapter $\Delta\theta_\text{SFT}$ is saved as the SFT checkpoint $\pi_{\text{SFT}}$.

%\vspace{-1em}
\subsection{Stage~2: Reward Model Training}
%\vspace{-0.4em}
\label{sec:rm}

The reward model $r_\phi:\,(p,y)\mapsto\mathbb{R}$ is constructed by replacing the causal language-modeling head of $\pi_{\text{SFT}}$ with a single linear projection layer. A second LoRA adapter is trained together with the reward head and the transformer body. The scoring layer is fully fine-tuned to ensure the scalar reward is properly learned. Using the preference dataset $\mathcal{D}_\text{RM}$, the Bradley--Terry loss~\cite{bradley1952rank} is minimized as defined in~(\ref{eq:rm}).
\begin{equation}
\resizebox{0.9\columnwidth}{!}{%
$ \mathcal{L}_\text{RM}(\phi) = -\mathbb{E}_{(p,y^{w},y^{l})\sim\mathcal{D}_{\text{RM}}} \Big[ \log \sigma \big( r_\phi(p,y^{w}) - r_\phi(p,y^{l}) \big) \Big] $%
}
\label{eq:rm}
\end{equation}
where $\sigma$ represents the logistic sigmoid function. Training is performed for two epochs using AdamW~\cite{loshchilov2017decoupled} with a learning rate of $10^{-5}$, a per-device batch size of 8, and a gradient accumulation step of 2. The training budget is limited to reduce the risk of generating brittle reward patterns. Due to the additive invariance of the Bradley--Terry loss, the reward difference $\Delta$ is reported separately for each configuration, while a shared reward model is used for comparisons across models.
%%%=====
%%%=====
%\vspace{-0.8em}
\subsection{Stage~3: PPO Fine-Tuning and Stabilization}
%\vspace{-0.4em}
\label{sec:ppo}

The policy $\pi_{\text{RL}}$ is optimized using the objective function defined in~(\ref{eq:ppo_obj}).
\begin{equation}
\resizebox{0.88\columnwidth}{!}{%
$J(\theta) = \mathbb{E}_{p\sim\mathcal{D},\,y\sim\pi_{\theta}(\cdot|p)} \Big[ r_\phi(p,y) - \beta\,\mathrm{KL}\big(\pi_{\theta}(\cdot|p) \, \Vert \, \pi_{\text{SFT}}(\cdot|p)\big) \Big]$%
}
\label{eq:ppo_obj}
\end{equation}
PPO~\cite{schulman2017proximal} is applied with GAE~\cite{schulman2015high} ($\lambda=0.95,\,\gamma=1.0$), a clipping ratio $\epsilon=0.2$, and an adaptive KL coefficient~\cite{ziegler2019fine} targeting a mean per-token KL of $6.0$ nats, implemented using TRL~\cite{vonwerra2022trl}. Following standard RLHF practice~\cite{ziegler2019fine, ouyang2022training}, the reverse (mode-seeking) KL divergence, $\mathrm{KL}(\pi_\theta\Vert\pi_{\text{SFT}})$, is used as a penalty term. The current policy is represented by $\pi_\theta$ and the frozen SFT prior is represented by $\pi_{\text{SFT}}$. The divergence is estimated on a per-token basis using sampled roll-outs. Three systemic instabilities specific to the SLM scale are addressed at this stage.

\paragraph{Failure Mode I--Gradient Flow Obstruction in PEFT}
In some PEFT implementations, adapter parameters are incorrectly set as non-trainable when used in the PPO training loop~\cite{vonwerra2022trl}. As a result, the policy generates roll-outs and computes the loss, but its underlying distribution remains unchanged because the gradient flow is blocked. 
This is reduced using a merge-and-reinitialize method mentioned in Algorithm~\ref{alg:merge}. 
The SFT adapter is merged into the base model weights using the weight-merging operation ($\text{Merge}(\cdot)$). Then, a new zero-initialized LoRA adapter is attached using ($\text{InitPolicy}(\cdot)$). The policy is initialized from the SFT distribution while keeping all required parameters trainable. The merged weights $\hat{\theta}_{\text{SFT}}$ are also used as the frozen reference policy $\pi_{\text{ref}}$, ensuring that $\pi_{\text{ref}} \equiv \pi_{\text{SFT}}$ throughout training.

\begin{algorithm}[!tb]
\scriptsize
\caption{Merge-and-Reinitialize for PEFT--PPO}
\label{alg:merge}
\begin{algorithmic}[1]
\Require SFT adapter $\mathcal{A}^{\text{ad}}_{\text{SFT}}$, base model $\theta_0$, output dir $\mathcal{O}$
\Ensure Trained policy $\pi_\theta$
\State $\hat{\theta}_{\text{SFT}} \gets \text{Merge}(\theta_0, \mathcal{A}^{\text{ad}}_{\text{SFT}})$; save to $\mathcal{O}/\text{merged}$ \Comment{fold SFT into base}
\State $\ell \gets \text{LoraConfig}(r,\alpha,\text{dropout}{=}0)$ \Comment{fresh, zero-init adapter}
\State $\pi_\theta \gets \text{InitPolicy}(\hat{\theta}_{\text{SFT}},\ell)$; $\pi_{\text{ref}} \gets \text{InitPolicy}(\hat{\theta}_{\text{SFT}})$
\State $\text{snapshot} \gets \text{copy of trainable parameters of } \pi_\theta$
\For{$t = 1$ \textbf{to} $T_{\text{PPO}}$}
    \State Sample $(p_b,y_b)_{b=1}^{B}$ from $\pi_\theta$; $r_b \gets r_\phi(p_b,y_b)$
    \State Whiten $r_b$, clip to $[-3,3]$, compute advantages via GAE
    \If{$\overline{\rho}_{\text{mini-batch}} > 5$} \textbf{skip} mini-batch \EndIf
    \State Update $\pi_\theta$ with PPO in \texttt{float32}
    \If{any parameter of $\pi_\theta$ is NaN or Inf}
        \State $\pi_\theta \gets \text{snapshot}$; reset optimizer moments \Comment{rollback}
    \Else
        \State $\text{snapshot} \gets \text{copy of trainable parameters of } \pi_\theta$
    \EndIf
\EndFor
\State \Return $\pi_\theta$
\end{algorithmic}
\end{algorithm}

\paragraph{Failure Mode II--Numerical Instability in Reduced Precision}
PPO importance ratios $\rho_t = \exp(\log\pi_\theta - \log\pi_{\text{SFT}})$, can suffer from catastrophic cancellation when computed in bfloat16 precision. This occurs because the difference between two similar log-probabilities loses numerical precision due to the limited seven-bit mantissa. In models with fewer than 200 million parameters, this precision limitation can cause probability-ratio values to exceed $10^6$ during the first optimization steps, resulting in hardware-level exceptions. This issue is addressed by using float32 precision for all tensors in the PPO loop, including the policy, reference model, value head, and reward model.

\paragraph{Failure Mode III--Distributional Collapse}
Long-tailed reward distributions combined with unclipped KL penalties can cause the policy to generate degenerate outputs. This occurs when the advantage estimates significantly exceed the clipping range $\epsilon$, causing the optimizer to move the policy toward regions where the reference model assigns very low probability. This mechanism includes reward whitening with $3\sigma$ clipping to limit the advantage estimates, an importance-ratio threshold of 5 to skip unstable mini-batches, and a weight-rollback mechanism. The weight-rollback mechanism saves the trainable parameters before each optimizer step and restores the previous state when NaN or Inf values are detected.

%%%%%──────────────────────────── End 3. Methodology───────────────────────────

% ─── 4. Experimental Setup ───────────────────────────────────────────────────
\vspace{-0.2em} %% for ArXiv only
\section{Experimental Setup}
\label{sec:setup}

%\vspace{-0.6em}
\subsection{Models}
%\vspace{-0.4em}
\label{sec:models}

Five small language models from two distinct architecture families were used to analyze the interaction between architecture and parameter count during RL-based alignment, as presented in Table~\ref{tab:models-and-data}. The selected models range from 70M to 410M parameters, providing broad coverage of model sizes within the SLM range. The Pythia family~\cite{biderman2023pythia} includes three deduplicated variants (70M, 160M, and 410M) that use a GPT-NeoX Byte-Pair Encoding (BPE) tokenizer and are pre-trained on the Pile dataset~\cite{gao2020pile}. The SmolLM2 family~\cite{allal2025smollm2} consists of 135M and 360M parameter models based on a Llama-style architecture with Rotary Positional Embeddings (RoPE), SwiGLU activations, and grouped-query attention. These models generate fifteen unique configuration pairs suitable for on-device deployment~\cite{liu2024mobilellm}.

%\vspace{-0.5em}
\subsection{Datasets}
%\vspace{-0.4em}
\label{sec:datasets}

Three corpora with different linguistic characteristics were selected: TinyStories ($\mathcal{D}_{\text{TS}}$)~\cite{eldan2023tinystories} for simple narratives, CNN/DailyMail ($\mathcal{D}_{\text{CNN}}$)~\cite{nallapati2016abstractive} for formal journalism, and Wikitext-103 ($\mathcal{D}_{\text{WT}}$)~\cite{merity2016pointer} for technical content. For each corpus, 10,000 examples were used for SFT. Preference pairs for reward-model training were generated using three degradation methods: truncation (cutting the chosen text in the middle of a clause), sentence shuffling (randomly changing the sentence order), and cross-example mismatch (using a continuation from a different reference text). This approach ensures a consistent and reproducible synthetic training signal. Table~\ref{tab:models-and-data} summarizes the model parameters and per-model SFT perplexity.

\begin{table}[!tb]
\centering
\caption{\textbf{Model parameters and per-corpus SFT perplexity (lower is better):} Each SFT checkpoint is a LoRA adapter trained on 10K samples for 5 epochs. \textbf{Bold} marks the best per-corpus value. SmolLM2 records the strongest SFT baseline across domains.}
\label{tab:models-and-data}
\footnotesize
\setlength{\tabcolsep}{3.5pt}
\begin{tabular}{llrr|ccc}
\toprule
\multirow{2}{*}{\textbf{Family}} & \multirow{2}{*}{\textbf{Model}} & \multirow{2}{*}{\textbf{Params}} & \multirow{2}{*}{\textbf{L}} & \multicolumn{3}{c}{\textbf{SFT PPL} $\downarrow$} \\
 & & & & TS & CNN & WT \\
\midrule
\multirow{3}{*}{Pythia~\cite{biderman2023pythia}}
  & Pythia-70M   & 70.4M  & 6  & 51.4  & 70.3  & 115.1 \\
  & Pythia-160M  & 162M   & 12 & 13.5  & 29.4  & 53.5  \\
  & Pythia-410M  & 405M   & 24 & 6.5   & 16.2  & 25.4  \\
\midrule
\multirow{2}{*}{SmolLM2~\cite{allal2025smollm2}}
  & SmolLM2-135M & 135M   & 30 & 7.0   & 18.8  & 24.4  \\
  & SmolLM2-360M & 361M   & 32 & \textbf{5.3}   & \textbf{12.7} & \textbf{16.7} \\
\bottomrule
\end{tabular}
\end{table}

%\vspace{-0.5em}
\subsection{Training Hyperparameters}
%\vspace{-0.4em}
A fixed set of hyperparameters was used across all configurations without model-specific tuning. Optimization was performed using AdamW~\cite{loshchilov2017decoupled}. For SFT, a learning rate of $2\times10^{-5}$ was used with a batch size of $8\times 4$, a 6\% linear warm-up, and cosine decay over 5 epochs. Reward model training used a learning rate of $1\times10^{-5}$ and a batch size of $8\times 2$ over 2 epochs. PPO optimization utilized a learning rate of $5\times10^{-6}$, a rollout batch size of 32, and a mini-batch size of 4 over 250 steps. The PPO configuration used a clipping range $\epsilon=0.2$, GAE parameters $\lambda=0.95$ and $\gamma=1.0$, and an adaptive KL penalty $\beta$ with a target of 6.0 nats. LoRA rank $r$ was adjusted according to model capacity: $r=8$ for Pythia-70M, $r=16$ for Pythia-160M and SmolLM2-135M, and $r=32$ for Pythia-410M and SmolLM2-360M. To ensure numerical stability, float32 precision was used for all tensors in the PPO loop, while bfloat16 precision was used for SFT and reward-model training.

%\vspace{-0.7em}
\subsection{Evaluation Metrics}
%\vspace{-0.4em}
\label{sec:metrics}

Evaluation was conducted using 200 held-out prompts for each configuration. Perplexity (PPL) was computed on the generated continuations, with padding tokens masked to avoid inflated scores~\cite{merity2016pointer}. The reward score is computed as the mean output of $r_\phi$ on the model-generated responses. The reward gain $\Delta$ is defined as the signed difference between the PPO and SFT reward scores, $\bar{r}_\text{PPO} - \bar{r}_\text{SFT}$. The win rate is estimated as the probability that a PPO response receives a higher score than an SFT response. It is computed as $\Phi(\Delta / \sqrt{\sigma_\text{PPO}^2 + \sigma_\text{SFT}^2})$, where $\Phi$ represents the standard normal Cumulative Distribution Function (CDF). Lexical diversity was evaluated using Distinct-1 and Distinct-2, while n-gram overlap was assessed using ROUGE-1 and ROUGE-L~\cite{lin2004rouge}. Due to the additive invariance of the Bradley--Terry loss, absolute reward scores are compared only within each configuration, while a shared reward model is used for comparisons across different models.

%\vspace{-0.7em}
\subsection{Computational Environment and Reproducibility}
%\vspace{-0.4em}
\label{sec:hardware}

All experiments were conducted using two NVIDIA RTX A6000 GPUs (48\,GB each, CUDA 12.1) and required approximately 16 GPU-hours. A post-hoc cross-verification script was used to verify all 339 scalar result fields against the raw evaluation outputs. An interactive verification application is also provided with the released code.

% ─── End 4. Experimental Setup ───────────────────────────────────────────────────

\begin{figure}[!tb]
\centering
% \includegraphics[width=\columnwidth]{./figure/fig3_sft_vs_ppo_scatter.pdf}
% (left bottom right top)
\includegraphics[width=\columnwidth, trim=2mm 2.9mm 2mm 2mm, clip]{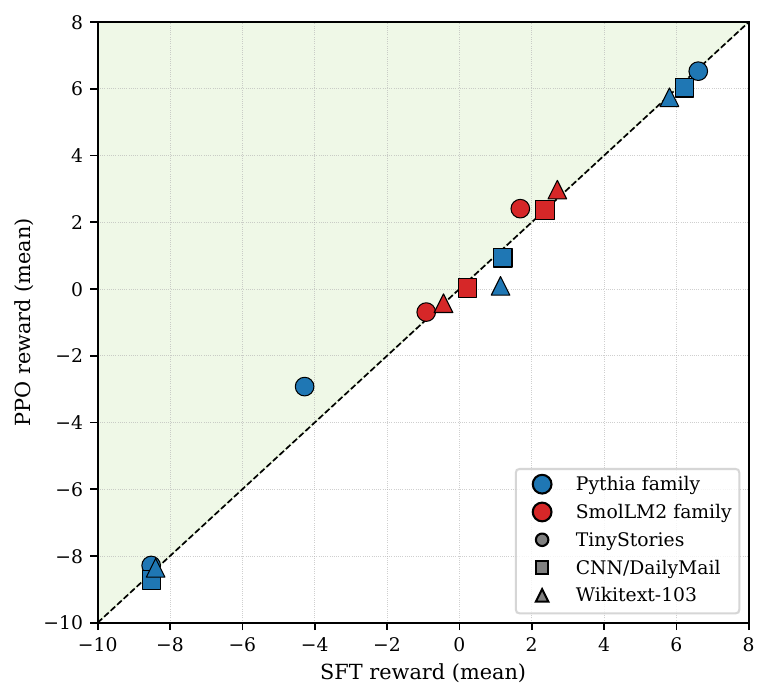}
%\vspace{-2.1em}
\caption{\textbf{SFT versus PPO reward across all 15 configurations}. Each marker represents the mean reward for one configuration, where color indicates the architecture family and shape represents the corpus. Markers above the dashed identity line indicate improvement by PPO, while the shaded upper-triangular region represents the ``PPO improves'' area. A clear rightward shift is observed for Pythia-410M and SmolLM2-360M on TinyStories, while Pythia-70M remains close to the identity line.}
\label{fig:sft-vs-ppo}
\end{figure}

\begin{figure}[t]
\centering
% \includegraphics[width=\columnwidth]{./figure/fig5_capacity_headroom.pdf}
% (left bottom right top)
\includegraphics[width=\columnwidth, trim=2mm 2.3mm 2mm 2mm, clip]{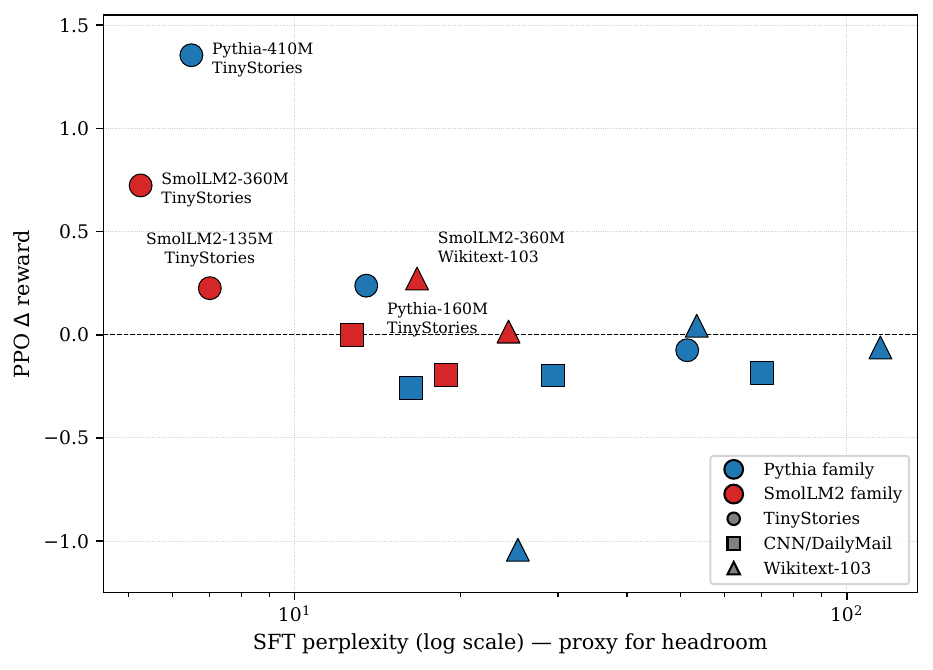}
%\vspace{-2.1em}
\caption{\textbf{Capacity-headroom hypothesis:} PPO gain is associated with a fluent SFT prior rather than raw parameter count. The figure shows SFT perplexity (log scale, $x$-axis) against PPO reward delta ($y$-axis) for all 15 configurations. A clear negative relationship is observed: PPO improves performance when the SFT prior is already fluent ($\text{PPL}\lesssim20$) but provides no improvement when the prior is weak.}
\label{fig:capacity-headroom}
\end{figure}

% ─── 5. Results ──────────────────────────────────────────────────────────────
\vspace{-0.5em} %% for ArXiv only
\section{Results Analysis}
\label{sec:results}
%\vspace{-0.6em}
\subsection{Main Performance Analysis}
%\vspace{-0.4em}
\label{sec:main-results}

The main results for all fifteen configurations are presented in Table~\ref{tab:main}. PPL and reward-model scores are reported for both the SFT and PPO checkpoints, together with the reward difference ($\Delta$) and analytical win rates. 
% %\vspace{-0.4em}
The largest positive reward gains are observed for Pythia-410M and SmolLM2-360M models on TinyStories ($\Delta\!=\!+1.355$ and $+0.724$, with win rates of 59.9\% and 59.7\%, respectively). On the other hand, Pythia-70M shows near-zero or negative changes across all domains and remains close to the identity diagonal in Fig.~\ref{fig:sft-vs-ppo}.

The reliability of these gains is assessed using a two-sided $z$-test on $n=200$ held-out prompts for each configuration, with a 95\% Confidence Interval (CI) reported for $\Delta$. Three configurations show statistically significant improvements: Pythia-410M on TinyStories (CI $[+0.61,+2.10]$, $p<0.001$), SmolLM2-360M on TinyStories (CI $[+0.32,+1.13]$, $p<0.001$), and SmolLM2-360M on Wikitext-103 (CI $[+0.04,+0.50]$, $p<0.05$). These results confirm that the observed gains are not due to sampling variation. The only significant regression is observed for Pythia-410M on Wikitext-103 ($\Delta=-1.043$, $p<0.001$). Its SFT prior has a PPL of 25.4, which is above the fluency threshold and supports the capacity-headroom hypothesis. The intervals are calculated from the complete per-prompt reward distribution rather than from repeated seeds, providing a direct measure of reliability within each configuration.
 
\begin{table*}[!tb]
\centering
\caption{\textbf{Main results:} Perplexity ($\downarrow$ is better) and reward-model score ($\uparrow$ is better) for SFT and PPO checkpoints across 5 architectures $\times$ 3 corpora. $\Delta = \bar{r}_\text{PPO} - \bar{r}_\text{SFT}$. Win\,\% is an analytical estimate $\Phi(\Delta/\sqrt{\sigma_\text{PPO}^2 + \sigma_\text{SFT}^2})$. Reward scores are reported on per-configuration scales; absolute values are not comparable across rows. \textbf{Bold} marks the best $\Delta$ and Win\,\% per family.}
% Best $\Delta$ and Win\,\% per family, and the best PPL per column per model family in \textbf{bold}.}
\label{tab:main}
\small
% \resizebox{\textwidth}{!}{
\setlength{\tabcolsep}{4pt}
\begin{tabular}{ll|cc|cccc}
\toprule
\multirow{2}{*}{\textbf{Model}} & \multirow{2}{*}{\textbf{Dataset}} &
  \multicolumn{2}{c|}{\textbf{SFT Baseline}} &
  \multicolumn{4}{c}{\textbf{PPO-Aligned}} \\
& & PPL $\downarrow$ & Reward $\uparrow$ & PPL $\downarrow$ & Reward $\uparrow$ & $\Delta$ $\uparrow$ & Win\,\% $\uparrow$ \\
\midrule
\multirow{3}{*}{Pythia-70M}
  & TinyStories   & 51.4  & $+$6.61\,$\pm$\,1.63 & 51.2  & $+$6.53\,$\pm$\,1.42 & $-$0.075 &  \textbf{48.6} \\
  & CNN/DailyMail & 70.3  & $+$6.22\,$\pm$\,1.21 & 70.5  & $+$6.04\,$\pm$\,1.23 & $-$0.187 & 45.7 \\
  & Wikitext-103  & 115.1 & $+$5.81\,$\pm$\,1.24 & 116.7 & $+$5.75\,$\pm$\,1.30 & \textbf{$-$0.062} &  \textbf{48.6} \\
\midrule
\multirow{3}{*}{Pythia-160M}
  & TinyStories   & 13.5  & $-$8.52\,$\pm$\,2.39 & 13.5  & $-$8.28\,$\pm$\,2.46 & \textbf{$+$0.238} & \textbf{52.8} \\
  & CNN/DailyMail & 29.4  & $-$8.52\,$\pm$\,1.31 & 29.4  & $-$8.71\,$\pm$\,1.19 & $-$0.198 & 45.6 \\
  & Wikitext-103  & 53.5  & $-$8.40\,$\pm$\,2.65 & 53.2  & $-$8.35\,$\pm$\,2.34 & $+$0.044 & 50.5 \\
\midrule
\multirow{3}{*}{Pythia-410M}
  & TinyStories   & 6.5   & $-$4.28\,$\pm$\,4.14 & 7.3   & $-$2.92\,$\pm$\,3.48 & \textbf{$+$1.355} & \textbf{59.9} \\
  & CNN/DailyMail & 16.2  & $+$1.20\,$\pm$\,1.76 & 17.1  & $+$0.94\,$\pm$\,1.79 & $-$0.259 & 45.9 \\
  & Wikitext-103  & 25.4  & $+$1.14\,$\pm$\,2.89 & 27.5  & $+$0.10\,$\pm$\,2.84 & $-$1.043 & 39.9 \\
\midrule
\multirow{3}{*}{SmolLM2-135M}
  & TinyStories   & 7.0   & $-$0.92\,$\pm$\,2.26 & 7.4   & $-$0.69\,$\pm$\,1.96 & \textbf{$+$0.226} & \textbf{53.0} \\
  & CNN/DailyMail & 18.8  & $+$0.22\,$\pm$\,1.90 & 19.2  & $+$0.03\,$\pm$\,1.90 & $-$0.194 & 47.1 \\
  & Wikitext-103  & 24.4  & $-$0.44\,$\pm$\,1.53 & 25.1  & $-$0.42\,$\pm$\,1.41 & $+$0.015 & 50.3 \\
\midrule
\multirow{3}{*}{SmolLM2-360M}
  & TinyStories   & 5.3  & $+$1.69\,$\pm$\,2.25 & 5.3   & $+$2.41\,$\pm$\,1.89 & \textbf{$+$0.724} & \textbf{59.7} \\
  & CNN/DailyMail & 12.7 & $+$2.36\,$\pm$\,1.09 & 12.8  & $+$2.36\,$\pm$\,1.05 & $-$0.001 & 50.0 \\
  & Wikitext-103  & 16.7  & $+$2.71\,$\pm$\,1.28 & 16.9  & $+$2.98\,$\pm$\,1.06 & $+$0.272 &56.5 \\
\bottomrule
\end{tabular}
% }
\end{table*}

% %\vspace{-1.1em}
% [REVIEWER-EDIT 2026-04-26] Replacement Table III with reconciled D1 values.
\begin{table*}[b]
\centering
% \caption{\textbf{Comparison with publicly released SOTA SLM baselines:} All models are scored with the same reward model per dataset. PPL is computed on reference text; R is mean reward-model score; D1 is distinct-1 diversity. The \emph{Proposed (SFT)} row uses only 5 epochs of LoRA SFT on 10K examples; the \emph{Proposed (PPO)} row adds the full RLHF pipeline. }
\caption{\textbf{Comparison with publicly available SOTA SLM baselines:}
All models are evaluated using the same reward model for each dataset. PPL is computed using reference text; R denotes the mean reward-model score; D1 is distinct-1 diversity. The \emph{proposed (SFT)} rows use only 5 epochs of LoRA SFT on 10K examples, while the \emph{proposed (PPO)} rows include the complete RLHF pipeline. \textbf{Bold} indicates the best value in each column within each parameter class, i.e., the lowest PPL and the highest R and D1.}
% \caption{\textbf{Comparison with publicly released SOTA SLM baselines:} All models are scored with the same reward model per dataset. PPL is computed on reference text; R is mean reward-model score; D1 is corpus-level distinct-1 diversity. The \emph{Proposed (SFT)} row uses only 5 epochs of LoRA SFT on 10K examples; the \emph{Proposed (PPO)} row adds the full RLHF pipeline. All numbers are reproducible from \texttt{results/all\_results.json}.}
\label{tab:sota}
% \small
\resizebox{\textwidth}{!}{
\setlength{\tabcolsep}{4pt}
\begin{tabular}{lll|ccc|ccc|ccc}
\toprule
\multirow{2}{*}{\textbf{Class}} & \multirow{2}{*}{\textbf{Model}} & \multirow{2}{*}{\textbf{Training Regime}} &
  \multicolumn{3}{c|}{\textbf{TinyStories}} & \multicolumn{3}{c|}{\textbf{CNN/DailyMail}} & \multicolumn{3}{c}{\textbf{Wikitext-103}} \\
& & & PPL$\downarrow$ & R$\uparrow$ & D1 & PPL$\downarrow$ & R$\uparrow$ & D1 & PPL$\downarrow$ & R$\uparrow$ & D1 \\
\midrule
\multirow{3}{*}{\shortstack[l]{135M class}}
  & SmolLM2-135M-Instruct~\cite{allal2025smollm2} & instruction-tuned (2T tok) & 8.5  & \textbf{$-$0.52} & \textbf{0.195} & 19.8 & \textbf{$+$0.35} & \textbf{0.283} & 34.3 & $-$0.79 & \textbf{0.297} \\
  & \textbf{SmolLM2-135M (proposed, SFT)}   & LoRA SFT, 5 ep, 10K ex  & \textbf{7.0} & $-$0.92 & 0.147 & \textbf{18.8} & $+$0.22 & 0.248 & \textbf{24.4} & $-$0.44 & 0.230 \\
  & \textbf{SmolLM2-135M (proposed, PPO)}   & + 250-step PPO RLHF     & 7.4 & $-$0.69 & 0.172 & 19.2 & $+$0.03 & 0.250 & 25.1 & \textbf{$-$0.42} & 0.269 \\
\midrule
\multirow{4}{*}{\shortstack[l]{360M class\\(incl.\ 500M)}}
  & SmolLM2-360M-Instruct~\cite{allal2025smollm2} & instruction-tuned (4T tok) & 6.6  & $+$1.35 & 0.220 & 14.7 & \textbf{$+$3.08} & \textbf{0.329} & 24.3 & $+$2.58 & \textbf{0.331} \\
  & Qwen2.5-0.5B-Instruct~\cite{qwen2025technical} & instruction-tuned (18T tok) & 7.2  & $+$1.32 & \textbf{0.229} & 19.9 & $+$2.58 & 0.245 & 25.8 & $+$1.83 & 0.282 \\
  & \textbf{SmolLM2-360M (proposed, SFT)}   & LoRA SFT, 5 ep, 10K ex  & \textbf{5.3} & $+$1.69 & 0.135 & \textbf{12.7} & $+$2.36 & 0.243 & \textbf{16.7} & $+$2.71 & 0.227 \\
  & \textbf{SmolLM2-360M (proposed, PPO)}   & + 250-step PPO RLHF     & \textbf{5.3} & \textbf{$+$2.41} & 0.156 & 12.8 & $+$2.36 & 0.247 & \ 16.9 & \textbf{$+$2.98} & 0.310 \\
\bottomrule
\end{tabular}
}
\end{table*}

%\vspace{-0.64em}
\subsection{Capacity-Headroom Hypothesis}
%\vspace{-0.4em}
Figure~\ref{fig:capacity-headroom} shows a monotonic negative relationship between SFT perplexity and PPO reward gain, with an inflection point at $\text{PPL}_{\text{SFT}} \approx 20$. This pattern supports the \emph{capacity-headroom hypothesis}, which suggests that PPO effectiveness depends on both a fluent SFT prior and a discriminative reward signal, rather than on parameter count alone. Under this framework, a fluent prior keeps the generated samples within the training distribution of the reward model. In contrast, a weak prior generates a gradient that is difficult to distinguish from noise, which explains the lack of policy improvement for Pythia-70M across all domains. The hypothesis is supported by the results in Table~\ref{tab:main}. The configurations with $\text{PPL}_{\text{SFT}}>20$ show negligible or negative reward gains, while all reward gains above $+0.2$ occur only below this threshold. This boundary is presented as an empirical decision rule based on the evaluated corpora and the 70--500M parameter range, rather than as a universal threshold.

%\vspace{-0.7em}
\subsection{Comparison with SOTA Instruction-Tuned Baselines}
%\vspace{-0.4em}
A comparison with publicly available instruction-tuned baselines is presented in Table~\ref{tab:sota}. Domain-specific SFT consistently achieves lower perplexity than SmolLM2-360M-Instruct and Qwen2.5-0.5B-Instruct on all tested datasets. At the 360M parameter scale, the PPO-aligned checkpoint achieves the highest reward scores on both TinyStories ($+2.41$) and Wikitext-103 ($+2.98$). On Wikitext-103, it also reduces most of the lexical-diversity gap with the instruction-tuned baselines ($0.310$ vs. Qwen $0.282$ and SmolLM2-360M-Instruct $0.331$), while using several orders of magnitude less training data.

%\vspace{-0.7em}
\subsection{Text Quality and Diversity}
%\vspace{-0.4em}
Text diversity and reference-overlap metrics are shown in Table~\ref{tab:quality}. PPO-aligned D-1 remains within $\pm 0.04$ of the SFT prior in fourteen of the fifteen configurations. The only exception is SmolLM2-360M on Wikitext-103, where D-1 increases by $+0.083$ together with the largest reward gain in this capacity class. Within the SmolLM2 family, D-1 improves across all configurations. In the Pythia family, a small reduction in D-1 is observed when the reward signal is uninformative, indicating a shift toward the SFT distribution rather than policy collapse. ROUGE-1 and ROUGE-L remain stable across most configurations. The largest reduction is observed for the two SmolLM2 models on Wikitext-103, where increased diversity is associated with lower reference overlap. No evidence of mode collapse is observed in any configuration. A representative example is shown in Table~\ref{tab:qual}, where the SFT continuation becomes repetitive, while the PPO continuation remains relevant to the topic.
% ROUGE and BLEU scores remain within reporting noise of the SFT baseline, and no mode-collapse signature is detected.

\begin{table}[!t]
\centering
\caption{Text diversity and reference-overlap metrics for the SFT prior and PPO-aligned policy across all fifteen configurations. \textbf{Bold} indicates improvement in the PPO policy compared with the SFT prior.}
% \caption{Text diversity (corpus-level Distinct-1/2) and reference overlap (ROUGE-1/L) for the SFT prior and the PPO-aligned policy across all fifteen (model, corpus) configurations. Bold indicates PPO improvement over the SFT prior on the same metric and row. Values are taken verbatim from \texttt{results/all\_results.json}.}
\label{tab:quality}
% \scriptsize
% \setlength{\tabcolsep}{3pt}
\resizebox{\columnwidth}{!}{%
\begin{tabular}{ll|cccc|cccc}
\toprule
\multirow{2}{*}{\textbf{Model}} & \multirow{2}{*}{\textbf{Dataset}} &
  \multicolumn{4}{c|}{\textbf{SFT Prior}} &
  \multicolumn{4}{c}{\textbf{PPO Policy}} \\
& & D-1 & D-2 & R-1 & R-L & D-1 & D-2 & R-1 & R-L \\
\midrule
\multirow{3}{*}{Pythia-70M}
  & TinyStories   & 0.051 & 0.123 & 0.172 & 0.143 & 0.050 & 0.120 & 0.166 & 0.138 \\
  & CNN/DailyMail & 0.130 & 0.329 & 0.184 & 0.115 & 0.124 & 0.310 & 0.178 & 0.111 \\
  & Wikitext-103      & 0.084 & 0.189 & 0.118 & 0.095 & 0.081 & 0.186 & 0.110 & 0.087 \\
\midrule
\multirow{3}{*}{Pythia-160M}
  & TinyStories   & 0.103 & 0.320 & 0.241 & 0.161 & \textbf{0.104} & \textbf{0.329} & \textbf{0.246} & \textbf{0.163} \\
  & CNN/DailyMail & 0.195 & 0.526 & 0.237 & 0.130 & \textbf{0.201} & \textbf{0.543} & 0.236 & \textbf{0.131} \\
  & Wikitext-103      & 0.143 & 0.399 & 0.171 & 0.119 & 0.132 & 0.366 & \textbf{0.172} & \textbf{0.120} \\
\midrule
\multirow{3}{*}{Pythia-410M}
  & TinyStories   & 0.119 & 0.456 & 0.351 & 0.202 & 0.110 & 0.390 & 0.323 & 0.194 \\
  & CNN/DailyMail & 0.225 & 0.632 & 0.263 & 0.139 & 0.213 & 0.609 & \textbf{0.274} & \textbf{0.145} \\
  & Wikitext-103      & 0.165 & 0.541 & 0.219 & 0.139 & 0.144 & 0.493 & 0.216 & \textbf{0.140} \\
\midrule
\multirow{3}{*}{SmolLM2-135M}
  & TinyStories   & 0.147 & 0.513 & 0.310 & 0.181 & \textbf{0.172} & \textbf{0.579} & 0.253 & 0.154 \\
  & CNN/DailyMail & 0.248 & 0.682 & 0.246 & 0.129 & \textbf{0.250} & 0.678 & 0.232 & 0.123 \\
  & Wikitext-103      & 0.230 & 0.638 & 0.185 & 0.120 & \textbf{0.269} & \textbf{0.675} & 0.149 & 0.102 \\
\midrule
\multirow{3}{*}{SmolLM2-360M}
  & TinyStories   & 0.135 & 0.489 & 0.355 & 0.212 & \textbf{0.156} & \textbf{0.516} & 0.335 & 0.205 \\
  & CNN/DailyMail & 0.243 & 0.666 & 0.256 & 0.136 & \textbf{0.247} & \textbf{0.670} & 0.249 & 0.133 \\
  & Wikitext-103      & 0.227 & 0.651 & 0.206 & 0.138 & \textbf{0.310} & \textbf{0.731} & 0.141 & 0.102 \\
\bottomrule
\end{tabular}%
}
\end{table}

\begin{table}[t]
\centering
% \caption{Qualitative comparison on TinyStories (Pythia-410M). Prompt from the evaluation set followed by truncated SFT and PPO continuations. The SFT continuation loops, whereas the PPO continuation maintains topical focus, consistent with the $+1.36$ reward delta and with the stabilisation framework preventing the degenerate output regimes commonly observed in small-scale RLHF.}
\caption{Qualitative comparison on TinyStories (Pythia-410M). A prompt from the evaluation set is followed by truncated SFT and PPO continuations. The SFT continuation becomes repetitive, while the PPO continuation maintains topical relevance.}
\label{tab:qual}
\footnotesize
% \scriptsize
\begin{tabular}{p{0.95\columnwidth}}
\toprule
\textbf{Prompt:} ``Once upon a time, there was a mama and her little child. The child was only three years old\ldots'' \\
\midrule
\textbf{SFT:} ``\ldots boy thought it was great, but he was very shy. So, the mama said, `no, the laundry must not be done here. Let's go and clean the house'. The little boy said, `I can't do the laundry. I am a little boy' \textit{[unresolved loop]}'' \\
\textbf{PPO:} ``\ldots boy was very excited and ran to the laundry machine. He ran to the clothes and washed them carefully. He did not know that he would be doing it quickly and carefully. The little boy was so excited to \textit{[continues on-topic]}'' \\
\midrule
\textit{Per-configuration rewards: SFT $-4.28$, PPO $-2.92$; $\Delta$ reward $+1.355$.} \\
\bottomrule
\end{tabular}
\end{table}

%\vspace{-0.7em}
\subsection{Ablation: Effect of the Stabilization Mechanism}
%\vspace{-0.4em}
The impact of the proposed engineering methods is evaluated through an ablation study using the Pythia-70M model on the TinyStories dataset. Three configurations are compared: (a) Naive PEFT, where the SFT adapter is directly passed to the trainer; (b) Manual unfreeze, where LoRA parameters are made trainable without using a reference model; and (c) the proposed stabilization mechanism. Configuration (a) produces a reward delta of zero because no gradient flows through the LoRA parameters. Configuration (b) fails during the first mini-batch due to numerical instability and the absence of a reference distribution. Only configuration (c) completes the 250-step training run without catastrophic failure. Each mechanism addresses a specific failure mode: merge-and-reinitialize restores gradient flow in (a), float32 precision with an importance-ratio cap ($\bar\rho\le 5$) prevents overflow in (b), and reward whitening with $3\sigma$ clipping and weight rollback prevents policy collapse, keeping all fifteen runs within $15\%$ of the SFT perplexity. 

% %% ─── End 5. Results and Analysis ──────────────────────────────────────────────────────

% ─── 6. Discussion ───────────────────────────────────────────────────────────
\vspace{-0.3em} %% for ArXiv only
\section{Discussion}
%\vspace{-0.4em}
\label{sec:discussion}

\noindent\textbf{Engineering Recommendations.}\ To ensure the stability of SLM-scale training, three essential methods are implemented. Merge-and-reinitialization is applied when TRL is combined with LoRA-adapted policies, while float32 precision is recommended for models with fewer than 200M parameters, and weight-rollback is used to prevent unrecoverable policy collapse. The convergence process is mainly controlled by an adaptive KL target of 6.0 nats and an importance-ratio cap of $\bar\rho \le 5$, where $\bar\rho$ is the mini-batch mean of $\rho_t$. The importance-ratio cap is the most sensitive parameter for models with fewer than 200M parameters. Since the loop remains stable across the tested learning-rate and clipping-ratio ranges, the same configuration is used for all fifteen cases. The merge operation is performed once before training with $\mathcal{O}(|\theta|)$ complexity and is independent of the number of PPO steps. At larger scales, the same gradient-flow property can be achieved by re-enabling the adapter gradients after the PEFT round-trip.

\noindent\textbf{Deployment Decision Rule.}\ The capacity-headroom hypothesis provides a practical guideline for SLM agents in adaptive and human–machine systems. A 70--500M parameter agent can be reliably aligned when (i) the supervised prior is sufficiently fluent ($\text{PPL}<20$) to keep samples within the training distribution of the reward model, and (ii) the reward signal is sufficiently discriminative to provide a useful gradient. When $\text{PPL}\in[20,50]$, the expected improvement is limited and regression may occur, as observed for Pythia-410M on Wikitext-103. In this range, computational resources may be better used to improve SFT data quality or increase the LoRA rank. When PPL exceeds $50$, PPO is unlikely to provide improvement. PPO may drift or collapse if either condition is not satisfied, while the safety mechanism can reduce but not fully resolve the underlying capacity limitation. These findings connect the empirical results with the broader concept of hybrid and cybernetically regulated computational-intelligence systems.

\noindent\textbf{Forward Compatibility with Multi-Turn Agents.}\ The empirical claims of this paper are limited to the single-turn form of the agent MDP, where the action space is reduced to the vocabulary $\mathcal{V}$, the horizon consists of one turn, and the reward is provided at the terminal state. A forward-compatibility framework for the multi-turn case is released with the fifteen checkpoints. It includes a parser for four sentinel-delimited action types, a set of deterministic and verifiable tools, and a multi-turn PPO variant that applies the safety mechanism at each turn. This framework is released as an unvalidated, forward-looking component rather than an evaluated contribution, and no multi-turn experiments are reported. The three identified failure modes remain necessary conditions for any form of the agent MDP, which motivates releasing the framework as a basis for future work on tool-call reward design and per-turn safety analysis.

\noindent\textbf{Limitations and Future Work.}\ Five boundary conditions are considered to define the scope of this study. First, synthetic preference pairs may limit the discriminative ability of the reward model, while human-annotated pairs could provide a stronger signal for complex corpora. Second, the 250-step PPO training budget is intentionally limited and may not capture the full improvements possible with longer training schedules. Third, the Bradley–Terry objective is additively invariant, which makes absolute reward values incomparable across different configurations. Fourth, the evaluation is limited to generative quality, while downstream task performance and human preference studies are needed for a complete evaluation of agent utility. Finally, the empirical evaluation is limited to the single-turn form of the agent MDP. The released forward-compatibility framework supports future extension of the safety mechanism to multi-turn interactions, where tool invocations and external observations increase the state space. Since DPO
does not use the explicit reward model and safety mechanism investigated in this paper, a controlled empirical comparison would address a different research question and is considered a direct extension of this work.

%% ─── END 6. Discussion ───────────────────────────────────────────────────────────

% ─── 7. Conclusion ───────────────────────────────────────────────────────────
\vspace{-0.4em} %% for ArXiv only
\section{Conclusion}
%\vspace{-0.4em}
\label{sec:conclusion}

This paper investigates the applicability of PPO for aligning small language model agents. Three reproducible failure modes at the 70--500M scale were identified: silent LoRA gradient freezing, bfloat16 importance-ratio overflow, and reward-driven policy collapse. These issues were addressed using a three-layer cybernetic safety mechanism that transforms the PPO update from an unstable process into a more reliable engineering approach. The resulting checkpoints achieve competitive performance compared with SOTA instruction-tuned baselines while using significantly less training data. The results support the capacity-headroom hypothesis across fifteen (model, corpus) configurations, showing that PPO effectiveness at the SLM scale depends on both a fluent SFT prior and a discriminative reward signal, rather than on the absolute number of model parameters. It remains an open research question whether this boundary also applies to group-relative methods such as GRPO and REINFORCE Leave-One-Out (RLOO), models at the 1B parameter scale, and multi-turn agent MDPs. The released checkpoints, datasets, scripts, and forward-compatibility framework are expected to facilitate these future investigations.

%%─── End 7. Conclusion ───────────────────────────────────────────────────────────

% %\vspace{0.5em}
% \section*{Acknowledgment}
% The computational resources used in this paper were provided by the Centre for Pattern Analysis and Machine Intelligence (CPAMI) Lab at the University of Waterloo. The authors also acknowledge the use of Google AI Studio and Claude (Anthropic) for linguistic refinement of the final manuscript to enhance its clarity.

%================= REFERENCES ===================================
\vspace{-0.3em} %% for ArXiv only
%%============================================
\balance
\bibliographystyle{IEEEtran} % Use the IEEE citation style

\bibliography{references}    % Load your .bib file (without the .bib extension)

@article{brown2020language,
  title={{Language Models are Few-Shot Learners}},
  author={Brown, Tom and Mann, Benjamin and Ryder, Nick and Subbiah, Melanie and Kaplan, Jared D and Dhariwal, Prafulla and Neelakantan, Arvind and Shyam, Pranav and Sastry, Girish and Askell, Amanda and others},
  journal={Advances in Neural Information Processing Systems},
  volume={33},
  pages={1877--1901},
  year={2020}
}

@article{eldan2023tinystories,
  title={{TinyStories: How Small Can Language Models Be and Still Speak Coherent English?}},
  author={Eldan, Ronen and Li, Yuanzhi},
  journal={arXiv:2305.07759},
  year={2023}
}

@inproceedings{biderman2023pythia,
  title={{Pythia: A Suite for Analyzing Large Language Models Across Training and Scaling}},
  author={Biderman, Stella and Schoelkopf, Hailey and Anthony, Quentin Gregory and Bradley, Herbie and O’Brien, Kyle and Hallahan, Eric and Khan, Mohammad Aflah and Purohit, Shivanshu and Prashanth, USVSN Sai and Raff, Edward and others},
  booktitle={International Conference on Machine Learning},
  pages={2397--2430},
  year={2023},
  organization={PMLR}
}

@article{allal2025smollm2,
  title={{SmolLM2: When Smol Goes Big--Data-Centric Training of a Small Language Model}},
  author={Allal, Loubna Ben and Lozhkov, Anton and Bakouch, Elie and Bl{\'a}zquez, Gabriel Mart{\'\i}n and Penedo, Guilherme and Tunstall, Lewis and Marafioti, Andr{\'e}s and Kydl{\'\i}{\v{c}}ek, Hynek and Lajar{\'\i}n, Agust{\'\i}n Piqueres and Srivastav, Vaibhav and others},
  journal={arXiv:2502.02737},
  year={2025}
}

@article{liu2024mobilellm,
    title={{MobileLLM: Optimizing Sub-billion Parameter Language Models for On-Device Use Cases}},
    author={Liu, Zechun and Zhao, Changsheng and Iandola, Forrest and Lai, Chen and Tian, Yuandong and Fedorov, Igor and Xiong, Yunyang and Chang, Ernie and Shi, Yangyang and Krishnamoorthi, Raghuraman and others},
    journal={arXiv:2402.14905},
    year={2024}
}

@article{qwen2025technical,
  title={{{Qwen2.5} Technical Report}},
  author={Yang, An and Yang, Baosong and Zhang, Beichen and Hui, Binyuan and Zheng, Bo and Yu, Bowen and Li, Chengyuan and Liu, Dayiheng and Huang, Fei and Wei, Haoran and others},
  journal={arXiv:2412.15115},
  year={2024}
}

@article{rafailov2023direct,
  title={{Direct Preference Optimization: Your Language Model is Secretly a Reward Model}},
  author={Rafailov, Rafael and Sharma, Archit and Mitchell, Eric and Manning, Christopher D and Ermon, Stefano and Finn, Chelsea},
  journal={Advances in Neural Information Processing Systems},
  volume={36},
  pages={53728--53741},
  year={2023}
}

@article{schulman2017proximal,
  title={{Proximal Policy Optimization Algorithms}},
  author={Schulman, John and Wolski, Filip and Dhariwal, Prafulla and Radford, Alec and Klimov, Oleg},
  journal={arXiv:1707.06347},
  year={2017}
}

@article{ouyang2022training,
  title={{Training Language Models to Follow Instructions with Human Feedback}},
  author={Ouyang, Long and Wu, Jeffrey and Jiang, Xu and Almeida, Diogo and Wainwright, Carroll and Mishkin, Pamela and Zhang, Chong and Agarwal, Sandhini and Slama, Katarina and Ray, Alex and others},
  journal={Advances in Neural Information Processing Systems},
  volume={35},
  pages={27730--27744},
  year={2022}
}

@article{christiano2017deep,
  title={{Deep Reinforcement Learning from Human Preferences}},
  author={Christiano, Paul F and Leike, Jan and Brown, Tom and Martic, Miljan and Legg, Shane and Amodei, Dario},
  journal={Advances in Neural Information Processing Systems},
  volume={30},
  year={2017}
}

@inproceedings{nallapati2016abstractive,
  title={{Abstractive Text Summarization Using Sequence-to-Sequence {RNNs} and Beyond}},
  author={Nallapati, Ramesh and Zhou, Bowen and dos Santos, Cicero and Gulcehre, Caglar and Xiang, Bing},
  booktitle={Proceedings of the SIGNLL Conference on Computational Natural Language Learning},
  pages={280--290},
  year={2016}
}

@article{merity2016pointer,
  title={{Pointer Sentinel Mixture Models}},
  author={Merity, Stephen and Xiong, Caiming and Bradbury, James and Socher, Richard},
  journal={arXiv:1609.07843},
  year={2016}
}

@article{vonwerra2022trl,
  title={{{TRL}: Transformer Reinforcement Learning}},
  author={von Werra, Leandro and Belkada, Younes and Tunstall, Lewis and Beeching, Edward and Thrush, Tristan and Lambert, Nathan and Huang, Shengding},
  journal={GitHub},
  year={2022},
  url={https://github.com/huggingface/trl}
}

@article{stiennon2020learning,
  title={{Learning to Summarize with Human Feedback}},
  author={Stiennon, Nisan and Ouyang, Long and Wu, Jeffrey and Ziegler, Daniel and Lowe, Ryan and Voss, Chelsea and Radford, Alec and Amodei, Dario and Christiano, Paul F},
  journal={Advances in Neural Information Processing Systems},
  volume={33},
  pages={3008--3021},
  year={2020}
}

@article{touvron2023llama,
  title={{Llama 2: Open Foundation and Fine-Tuned Chat Models}},
  author={Touvron, Hugo and Martin, Louis and Stone, Kevin and Albert, Peter and Almahairi, Amjad and Babaei, Yasmine and Bashlykov, Nikolay and Batra, Soumya and Bhargava, Prajjwal and Bhosale, Shruti and others},
  journal={arXiv:2307.09288},
  year={2023}
}

@article{ethayarajh2024kto,
  title={{{KTO}: Model Alignment as Prospect Theoretic Optimization}},
    author={Ethayarajh, Kawin and Xu, Winnie and Muennighoff, Niklas and Jurafsky, Dan and Kiela, Douwe},
  journal={arXiv:2402.01306},
  year={2024}
}

@article{shao2024deepseekmath,
  title={{{DeepSeekMath}: Pushing the Limits of Mathematical Reasoning in Open Language Models}},
  author={Shao, Zhihong and Wang, Peiyi and Zhu, Qihao and Xu, Runxin and Song, Junxiao and Zhang, Mingchuan and Li, YK and Wu, Y and Guo, Daya},
  journal={arXiv:2402.03300},
  year={2024}
}

@article{hu2022lora,
  title={{LoRA: Low-Rank Adaptation of Large Language Models}},
  author={Hu, Edward J and Shen, Yelong and Wallis, Phillip and Allen-Zhu, Zeyuan and Li, Yuanzhi and Wang, Shean and Wang, Liang and Chen, Weizhu and others},
  journal={International Conference on Learning Representations (ICLR)},
  volume={1},
  number={2},
  pages={3},
  year={2022}
}

@article{dettmers2023qlora,
  title={{{QLoRA}: Efficient Finetuning of Quantized {LLMs}}},
  author={Dettmers, Tim and Pagnoni, Artidoro and Holtzman, Ari and Zettlemoyer, Luke},
  journal={Advances in Neural Information Processing Systems},
  volume={36},
  pages={10088--10115},
  year={2023}
}

@article{ziegler2019fine,
  title={{Fine-Tuning Language Models from Human Preferences}},
  author={Ziegler, Daniel M and Stiennon, Nisan and Wu, Jeffrey and Brown, Tom B and Radford, Alec and Amodei, Dario and Christiano, Paul and Irving, Geoffrey},
  journal={arXiv:1909.08593},
  year={2019}
}

@article{schulman2015high,
  title={{High-Dimensional Continuous Control Using Generalized Advantage Estimation}},
  author={Schulman, John and Moritz, Philipp and Levine, Sergey and Jordan, Michael and Abbeel, Pieter},
  journal={arXiv:1506.02438},
  year={2015}
}

@inproceedings{jiang2023neftune,
  title={{{NEFTune}: Noisy Embeddings Improve Instruction Finetuning}},
  author={Jain, Neel and Chiang, Ping-yeh and Wen, Yuxin and Kirchenbauer, John and Chu, Hong-Min and Somepalli, Gowthami and Bartoldson, Brian and Kailkhura, Bhavya and Schwarzschild, Avi and Saha, Aniruddha and others},
  booktitle={International Conference on Learning Representations},
  volume={2024},
  pages={17566--17591},
  year={2024}
}

@article{gao2020pile,
  title={{The Pile: An 800gb Dataset of Diverse Text for Language Modeling}},
  author={Gao, Leo and Biderman, Stella and Black, Sid and Golding, Laurence and Hoppe, Travis and Foster, Charles and Phang, Jason and He, Horace and Thite, Anish and Nabeshima, Noa and others},
  journal={arXiv:2101.00027},
  year={2020}
}

@article{loshchilov2017decoupled,
  title={{Decoupled weight decay regularization}},
  author={Loshchilov, Ilya and Hutter, Frank},
  journal={arXiv:1711.05101},
  year={2017}
}

@article{bradley1952rank,
  title={{Rank Analysis of Incomplete Block Designs: {I.} The Method of Paired Comparisons}},
  author={Bradley, Ralph Allan and Terry, Milton E},
  journal={Biometrika},
  volume={39},
  number={3/4},
  pages={324--345},
  year={1952}
}

@inproceedings{lin2004rouge,
  title={{{ROUGE}: A Package for Automatic Evaluation of Summaries}},
  author={Lin, Chin-Yew},
  booktitle={Text Summarization Branches Out},
  pages={74--81},
  year={2004}
}

%%==============================================
% \bibliography{references,Bibliography}

%%==================================================================

\end{document}